\documentclass{article}
\usepackage{spconf,amsmath,graphicx}
\usepackage{hyperref}
\usepackage{cleveref}
\usepackage{subfigure}
\usepackage{pdfpages}

\newcommand{\etal}{\textit{et al}. }

\title{AUTOMATIC PULMONARY LOBE SEGMENTATION USING DEEP LEARNING}
\name{Hao Tang\sthanks{These authors have contributed equally to this work.}$^{\dagger}$$^{\ddagger}$, Chupeng Zhang\footnotemark[1]$^{\dagger}$$^{\ddagger}$, Xiaohui Xie$^{\dagger}$}
\address{$^{\dagger}$Department of Computer Science, University of California, Irvine \\$^{\ddagger}$Deep Voxel Inc., 3200 Park Center Dr, Costa Mesa}
\begin{document}
%
\maketitle
\begin{abstract}
Pulmonary lobe segmentation is an important task for pulmonary disease related Computer Aided Diagnosis systems (CADs). Classical methods for lobe segmentation rely on successful detection of fissures and other anatomical information such as the location of blood vessels and airways. With the success of deep learning in recent years, Deep Convolutional Neural Network (DCNN) has been widely applied to analyze medical images like Computed Tomography (CT) and Magnetic Resonance Imaging (MRI), which, however, requires a large number of ground truth annotations. In this work, we release our manually labeled 50 CT scans which are randomly chosen from the LUNA16 dataset and explore the use of deep learning on this task. We propose pre-processing CT image by cropping region that is covered by the convex hull of the lungs in order to mitigate the influence of noise from outside the lungs. Moreover, we use a hybrid loss function with dice loss to tackle extreme class imbalance issue and focal loss to force model to focus on voxels that are hard to be discriminated. To validate the robustness and performance of our proposed framework trained with a small number of training examples, we further tested our model on CT scans from an independent dataset. Experimental results show the robustness of the proposed approach, which consistently improves performance across different datasets by a maximum of $5.87\%$ as compared to a baseline model.
The annotations are public available \url{https://github.com/deep-voxel/automatic_pulmonary_lobe_segmentation_using_deep_learning/} and are for non-commercial use only.
\end{abstract}
\section{Introduction}
\label{sec:intro}
Lung cancer has been the leading cause of all cancer-related disease during the past years \cite{siegel2015cancer}. Segmentation of pulmonary lobe based on Computed Tomography (CT) is an important task for Computer Aided Diagnosis systems (CADs). Pulmonary lobe segmentation is relevant in a wide range of clinical applications. The location and distribution of pulmonary disease can be a significant factor in determining the most suitable treatment. According to \cite{national2001patients}, locally distributed emphysema can be treated more effectively by lobar volume resection than homogeneously distributed emphysema. Another application is pulmonary nodule detection where detecting pulmonary nodule in its early stage is critical for a good prognosis of the disease. Recent success in deep learning especially the use of Deep Convolutional Neural Network (DCNN) has accelerated the development of automatic pulmonary nodule detection and classification system, such as \cite{tang2018automated, DBLP:journals/corr/abs-1709-05538}, which can be used to help radiologist and reduce their labor work. Precise segmentation of lung lobes can be used to generate automatic electronic diagnosis report since the rough location of nodules are required and the precise coordinate information is rarely used in most medical institutes.

Human lungs are composed of five lobes (two in the left lung and three in the right lung). The upper lobe and lower lobe of left lung are separated by the major fissures (oblique fissure). In the right lung, there are three lobes, namely upper lobe, middle lobe and lower lobe. The upper lobe and middle lobe are divided by the minor fissure (horizontal fissure) while the major fissure (oblique fissure) separates the lower lobe from the rest of the lung. 

Methods for pulmonary lobe segmentation have been focused on unsupervised models using classical computer vision techniques which usually include detecting fissures, locating bronchi and vessels, such as \cite{kuhnigk2005new, lassen2013automatic}. More recently, FJS Bragman \etal applied probabilistic model in enhanced fissure detection using fissure prior which yields accurate results under various fissure incompleteness. However, the attempt of using deep learning in this task is still rare \cite{8489677} because of the need for a large number of annotated training examples. Moreover, publicly available annotations for pulmonary lobe segmentation can hardly be found for supervised training of deep neural network. 

In this work, we collaborate with our radiologist on manually creating and releasing reference annotations from a randomly chosen subset from the LUNA16 \cite{setio2017validation}. Next, we present a framework using DCNN that can be trained effectively and robustly with a small number of training examples. In order to further validate the generalization ability and robustness of the trained deep neural net, we annotated 10 more CT scans from Tianchi dataset as a holdout test set. Experimental results show the proposed framework generalize well to CT scans collected from different sources, which yields a maximum of 5.87\% improvement as compared to a baseline model.

Our contributions of this work are summarized as below:

$a).$ We propose pre-processing CT image by cropping region that is covered by the convex hull of the lungs in order to mitigate the influence of noise outside the lungs. We use a hybrid loss function with dice loss to tackle extreme class imbalance issue and focal loss to force model to focus on voxels that are hard to be discriminated, similar to \cite{zhu2019anatomynet}. This achieves the state-of-art averaged dice coefficient of $91.48\%$ on the LUNA16 test set and $94.17\%$ on the Tianchi test set respectively.

$b).$ We release our reference annotations on 50 CT scans randomly chosen from LUNA16 for supervised pulmonary lobe segmentation study, which is the first publicly available dataset with reference annotations on this task.


\section{DATA}
\label{sec:dataset}

\subsection{Data and annotation}
We randomly chose 50 CT scans from LUNA16 \cite{setio2017validation} and collaborated with our radiologist in creating annotations for each CT scan. LUNA16 is a subset of LIDC-IDRI \cite{armato2011lung} for pulmonary nodules, which is the largest publicly available lung image dataset. LUNA16 was then created by removing CT scan that has a slice thickness greater than 3mm, inconsistent slice spacing or missing slices from LIDC-IDRI dataset to provide an evaluation framework for pulmonary nodule detection. LIDC-IDRI data uses the Creative Commons Attribution 3.0 Unported License.

Reference annotation for each CT scan was then manually delineated by radiologist using Chest Image Platform\footnote{\url{https://chestimagingplatform.org/about}}. This software platform is built on top of the 3D slicer and uses an interactive algorithm to perform lobe segmentation where user is required to mark points on three fissures \cite{kuhnigk2005new, lassen2013automatic}.

We used 40 of the annotated CT scans for training our model and 10 for testing on the LUNA16 dataset. In order to validate the robustness of our algorithm, we further annotated another 10 randomly chosen CT scans collected from a different source: Tianchi\footnote{\url{https://tianchi.aliyun.com/competition/rankingList.htm?raceId=231601&season=0}}, which is also a large-scaled competition dataset.

\subsection{Reference annotation availability}
50 annotations created on the LUNA16 dataset are publicly available for supervised lung lobe segmentation study. However, the 10 annotations made on the Tianchi dataset will not be made publicly available at this time.

\section{METHOD}
\label{sec:method}
We present in this section the framework using the deep convolutional neural network which includes pre-processing of removing regions outside the lung region, model architecture and hybrid loss function used to train the network.
\subsection{Pre-processing}
We propose pre-processing by cropping region covered by the convex hull of the lungs, which removes noise from different CT scans outside the lungs, as well as reducing the cost of GPU memory as the input volume is substantially smaller.

We start with normalizing the whole CT scan by truncating Hounsfield Unit (HU) values outside the range of $[-1000, 600]$. Next, we use OTSU to binarize the CT image. A binary morphology close is then used to remove regions outside the lungs and binary hole filling is applied to fill small isolated regions in the lung on a per slice base. The convex hull of the two lungs is computed and a binary morphology dilation using $5*5$ kernel is applied to preserve information near the border slice by slice.

\subsection{Model architecture}

\begin{figure}
\centering
\includegraphics[width=0.48\textwidth]{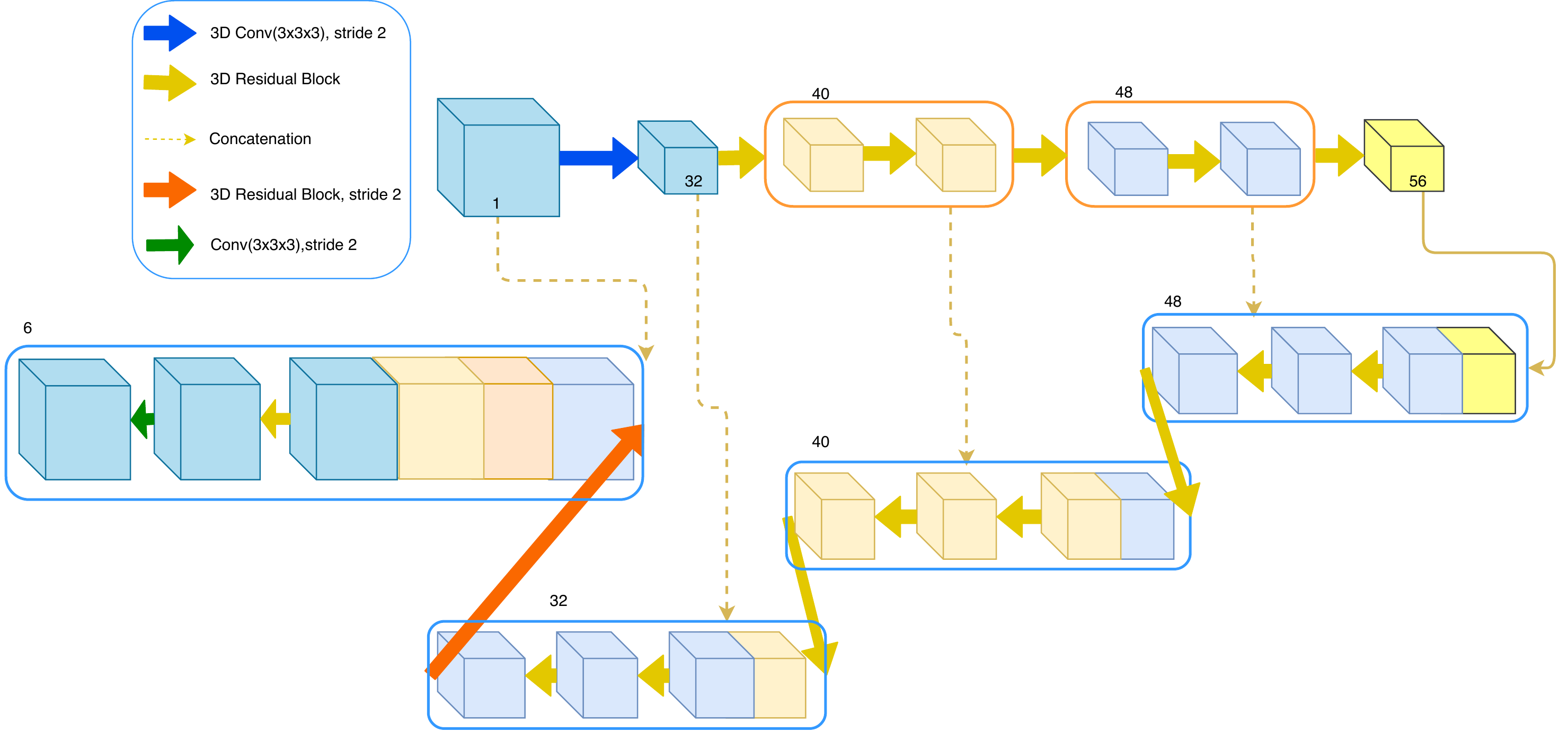}
\caption{Neural network architecture. Each cube represents 3D image volume and the side number denotes the number of channels in that block. Different from original V-Net, we only use one down-sampling to balance the trade-off between feature representation capacity and GPU memory.}
\label{fig:model} 
\end{figure} 

The network architecture is illustrated in \Cref{fig:model}. We use 3D residual block \cite{DBLP:journals/corr/HeZRS15} as a basic building block which consists of two consecutive $3*3*3$ convolution layers followed by ReLU and Batchnorm. We only use one down sampling in this architecture to balance the trade-off between feature representation capacity and GPU memory, which is different from original V-Net \cite{DBLP:journals/corr/MilletariNA16} who employs the standard four down-samplings. 

\subsection{Loss}
Dice loss is widely used for training a segmentation network using deep learning in the medical image. Dice loss performs relatively well when training samples are highly imbalanced as compared to cross entropy loss. However, dice loss fails to capture the difference of difficulty in classifying different voxels. For instance, voxels on the border are more difficult to be classified correctly than voxels are in the center of the lobe. As a result, we use a hybrid loss of both dice loss and focal loss \cite{8237586} to address voxel-wise imbalance and force model to focus on those voxels that are hard to be correctly predicted, similar to \cite{zhu2019anatomynet}:
\begin{equation}
\label{eq_loss}
L = L_{dice} + \lambda L_{focal} 
\end{equation}
\begin{equation*}
\label{eq_loss}
L_{dice} = \sum_c^C\sum_i^N{\frac{p_{ic} * g_{ic}}{p_{ic} * g_{ic} + (1 - p_{ic}) * g_{ic} + p_{ic} * (1 - g_{ic})}}
\end{equation*}
\begin{equation*}
\label{eq_loss}
L_{focal} = - \frac{1}{N}\sum_c^C\sum_i^N{\alpha_c * g_{ic} * (1 - p_{ic})^{\gamma} * log(p_{ic})}
\end{equation*}
$\lambda$ is a hyper-parameter controlling the balance between dice loss and focal loss, which is set to 1 in this work. $N$ is the total number of voxels in each mini-batch and $i$ is the index of each individual voxel. $C$ denotes the total number of classes which is six in this task (one more class for background). $p_{ic}$ is the predicted probability that $i$-th voxel is class $c$ and $g_{ic}$ is 1 if $i$-th voxel is class $c$ and 0 otherwise. $\alpha$ and $\gamma$ are parameters controlling weight for each class and adjusting the down-weighting of well-classified voxels respectively. We set $\alpha$ to be one and $\gamma$ to be two as suggested in \cite{8237586}.

\subsection{Data augmentation}
Data augmentation is critical for training model that can generalize well across different datasets, especially when the number of training samples is small. The input volume is randomly shifted, z-axis flipped and XY-plane rotated in order to improve the generalization ability of the model.

\section{EXPERIMENTS}
\label{sec:experiment}
Dice coefficient was used to evaluate the performance of the model:
\begin{equation}
\label{eq:dice_coefficient}
\begin{split}
DC_c(P, G)=2*\frac{P_c\cup G_c}{P_c\cap G_c}, c\in C
\end{split}
\end{equation}
\begin{equation}
\label{eq:avg_dice_coefficient}
\begin{split}
DC_{avg}(P, G)=\frac{1}{C}\sum_c^C DC_c(P_c, G_c)
\end{split}
\end{equation}
where $P$ is the set of predictions for each voxel and $G$ represents the set of ground truth label. We calculate dice coefficient for each lobe independently and averaged dice coefficient for all lobes as described in \Cref{eq:dice_coefficient} and \Cref{eq:avg_dice_coefficient} respectively.

We split 50 annotated CT scans from LUNA16 into 40 for training and 10 for testing. We tested our model on a holdout 10 CT scans annotated from Tianchi dataset as well to illustrate the robustness of our proposed approach. 

In order to assess the influence of hybrid loss and pre-processing by using convex hull, we added each component step by step and all models were trained using the same data augmentation with the same hyper parameters. We trained each model for 300 epochs using Adam as the optimizer and used the last epoch for predicting each test set. Batch size was set to one since the size of input volume might be different for each CT scan after pre-processing.

The step-wise gains of using hybrid loss and pre-processing by using convex hull are shown in \Cref{table:performance}. Also, we compared our best model with \cite{8489677} which we found most relative to our work. As shown in the tables, model trained with hybrid loss increased the averaged dice coefficient of the baseline model by 3.87\% on LUNA16 test set and 3.95\% on Tianchi test set. By removing regions outside the lung using convex hull further increased the averaged dice coefficient by 0.54\% on LUNA16 test set and 1.92\% on Tianchi test set as compared to the model trained only with hybrid loss. Moreover, the comparison between \cite{8489677} and our approach further validated the robustness and generalization ability of the proposed framework by improving the previous state-of-art result by a maximum of 2.39\% on averaged dice coefficient.

\begin{figure}
\centering
\includegraphics[width=0.48\textwidth]{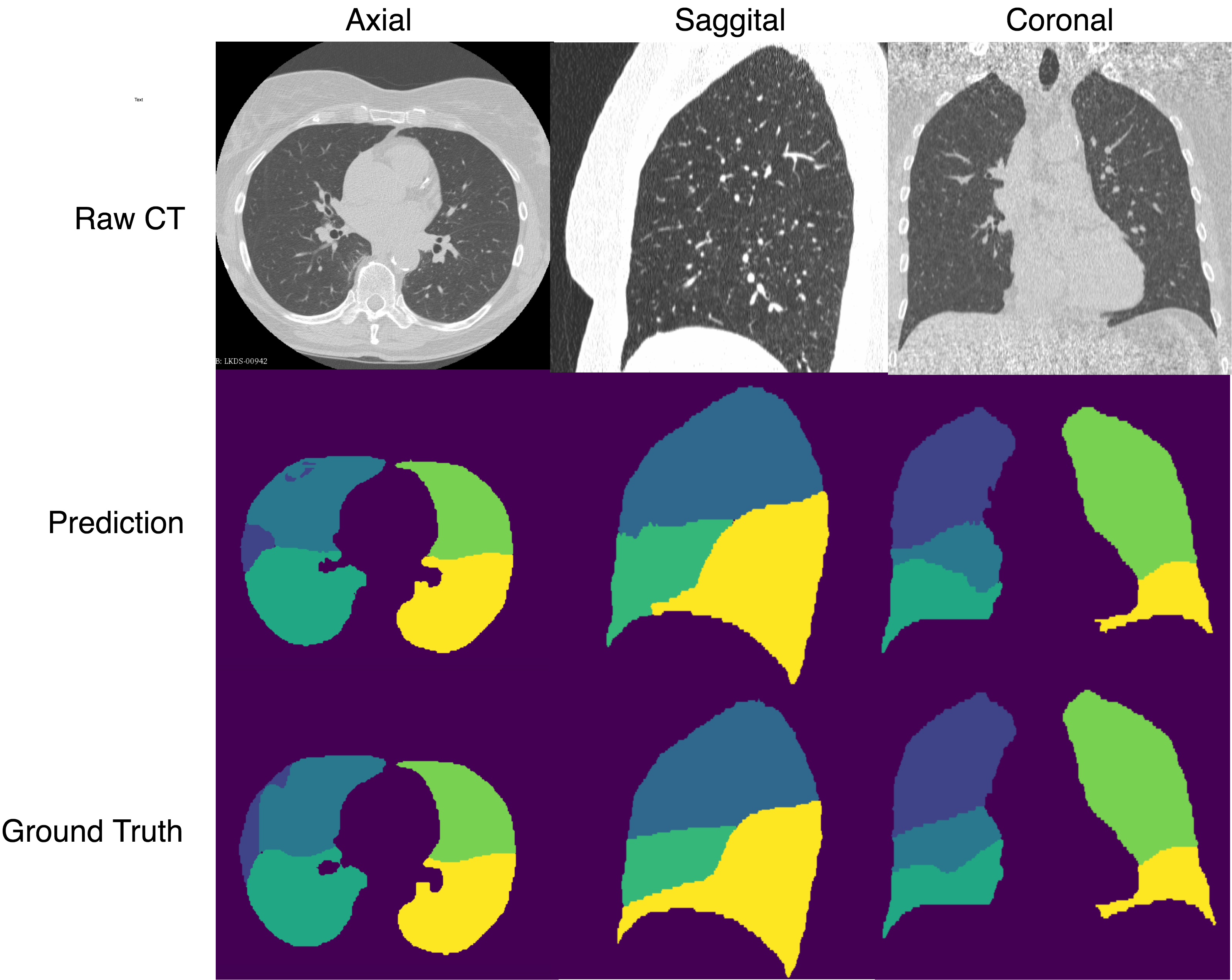}
\caption{Comparison between prediction of the model and ground-truth in CT scan views: Axial, Sagittal and Coronal.}
\label{fig:visualization} 
\end{figure}

We visualized in \Cref{fig:visualization} qualitative comparison between model prediction and reference annotation from three views, which illustrates the significance of focusing on hard negative examples and only regions inside and between two lungs. 


\begin{table}[]
\centering

\begin{tabular}{l r r r r r r}
\hline
\multicolumn{7}{c}{LUNA16 test set} \\ \hline \hline
      & RU & RM & RL & LU & LL & AVG \\ \hline
\cite{8489677}   & $\textbf{92.76}$ & $\textbf{84.68}$ & $\textbf{94.33}$ & $88.10$ & $94.78$  & $90.93$ \\ \hline
DL    & $78.28$ & $79.69$ & $93.96$ & $88.65$ & $94.79$  & $87.07$ \\ \hline
+ FL  & $90.58$ & $78.41$ & $93.95$ & $96.01$ & $95.77$  & $90.94$ \\ \hline
+ CH  & $92.53$ & $80.60$ & $93.05$ & $\textbf{96.10}$ & $\textbf{95.30}$  & $\textbf{91.48}$ \\ \hline
\end{tabular}
\begin{tabular}{l r r r r r r}
\hline
\multicolumn{7}{c}{Tianchi test set} \\ \hline \hline
      & RU & RM & RL & LU & LL & AVG \\ \hline
\cite{8489677}   & $93.11$ & $86.43$ & $94.54$ & $89.30$ & $95.40$  & $91.76$ \\ \hline
DL    & $80.80$ & $82.71$ & $94.46$ & $89.03$ & $94.51$  & $88.30$ \\ \hline
+ FL  & $92.59$ & $84.75$ & $93.08$ & $95.94$ & $94.88$  & $92.25$ \\ \hline
+ CH  & $\textbf{95.11}$ & $\textbf{87.92}$ & $\textbf{95.15}$ & $\textbf{97.21}$ & $\textbf{95.46}$  & $\textbf{94.17}$ \\ \hline
\end{tabular}
\caption{Step-wise performance gains of using hybrid loss and pre-processing using convex hull as compared to a baseline model trained only with dice loss and previous state-of-art method \cite{8489677}. RU, RM, RL, LU, LL and AVG represent the dice coefficient of right upper lobe, right middle lobe, right lower lobe, left upper lobe, left lower lobe and their average respectively. DL means model trained with dice loss and +FL means model trained with hybrid loss of focal loss and dice loss. +CH represents model trained with hybrid loss and training data cropped by the convex hull of the lungs.}
\label{table:performance}
\end{table}

\section{CONCLUSION}
\label{sec:conclusion}
In this work, we release our manual annotation by radiologist for 50 CT scans collected from the LUNA16 challenge and present a practical and robust framework for robust pulmonary lobe segmentation. We believe the public availability of those reference annotations will help the study of pulmonary lobe segmentation using supervised learning. Also, our proposed framework trained with a small number of training examples is proved to perform well across CT scans from different sources.

\bibliographystyle{IEEEbib}
\bibliography{refs}

\end{document}